# Deep Learning Approach to Anomaly Detection in Enterprise ETL Processes with Autoencoders

Xin Chen
Cornell University
Ithaca, USA

Saili Uday Gadgil
Virginia polytechnic institute and state University
Blacksburg, USA

Kangning Gao
The George Washington University
Washington, D.C, USA

Yi Hu
University of Southern California
Los Angeles, USA

Cong Nie*
Washington University in St. Louis
St. Louis, USA

*Abstract-In this paper, an anomaly detection method based on deep autoencoders is proposed to address anomalies that often occur in enterprise-level ETL data streams. The study first analyzes multiple types of anomalies in ETL processes, including delays, missing values, duplicate loading, and sudden abnormal changes, and applies data standardization and feature modeling to ensure stable and usable inputs. In the method design, the encoder-decoder structure compresses high-dimensional inputs into latent representations and reconstructs them, while reconstruction error is used to measure anomaly levels. Regularization constraints are introduced in the latent space to enhance feature sparsity and distribution learning, thereby improving robustness in complex data streams. Systematic analyses under different hyperparameter settings, environmental changes, and data characteristics show that the proposed method achieves superior performance in AUC, ACC, Precision, and Recall. The results demonstrate that the deep autoencoder-based detection mechanism can effectively capture latent distribution patterns in enterprise-level ETL data streams and accurately identify diverse anomalies, providing reliable support for enterprise data processing and intelligent analysis.*

*Keywords: Deep autoencoders; anomaly detection; ETL data flow; data governance*

## I. INTRODUCTION

In today's data-driven business environment, enterprises handle massive amounts of heterogeneous data every day. This data must pass through complex ETL (Extract, Transform, Load) processes, where it is extracted, transformed, and loaded into data warehouses or analytics platforms to support decision-making and intelligent analysis. However, ETL processes face many challenges[1]. These include dynamic changes in data sources, delays and losses during transmission, complexity of transformation rules, and storage pressure during the loading stage [2]. Such issues increase system uncertainty and create risks of anomalies. Once anomalies occur in ETL data flows, such as missing data, duplicate loading, inconsistent formats, or sudden abnormal changes, they can directly affect business operations and the accuracy of data analysis, and may even cause major deviations in decision outcomes. Therefore, monitoring and detecting anomalies in enterprise-level ETL data flows has become a key task to ensure data quality and business continuity[3].

Traditional ETL anomaly detection methods often rely on rule matching, statistical thresholds, or manually defined monitoring indicators. These approaches can be effective when dealing with structured and relatively stable data. Yet, with the acceleration of enterprise digital transformation, data sources have become highly complex and dynamic. A single rule is insufficient to handle the diverse anomaly patterns in such environments. For high-frequency and large-scale real-time data streams, traditional methods struggle to achieve both accuracy and efficiency. They are prone to missed detections and false alarms. At the same time, anomalies often have nonlinear and multi-dimensional characteristics. Their patterns may be hidden within the underlying data distribution, beyond the descriptive capacity of conventional techniques. In this context, there is an urgent need for more intelligent and automated detection methods that can capture complex anomaly patterns and ensure the stability and reliability of enterprise-level ETL data flows.

The rise of deep learning has created new opportunities for anomaly detection [4]. Among these methods, autoencoders have become an important tool for handling complex high-dimensional data due to their strong representation learning ability [5-7]. By compressing and reconstructing input data in unsupervised or weakly supervised settings, autoencoders can learn the underlying data distribution and feature space. Anomalous signals can then be reflected in reconstruction errors. This property is particularly suitable for anomaly detection in ETL data flows. In most enterprise scenarios, normal data greatly outnumbers anomalous data [8-10]. Autoencoders can learn normal patterns and highlight deviations from them. Compared with rule-based methods, deep autoencoder-based approaches adapt better to complex and dynamic environments [11]. They improve detection accuracy, reduce manual intervention costs, and show strong generalization and application potential.

From the perspective of enterprise applications, building an efficient ETL anomaly detection system is not only a technical issue. It is also a core element of data governance and business continuity management. As enterprises become increasingly dependent on data, data quality problems directly affect business outcomes [12]. In mild cases, they distort operational indicators. In severe cases, they lead to decision errors and economic losses. In industries such as finance, healthcare, retail, and telecommunications, data-driven real-time analysis and prediction are key to competitiveness. Anomaly detection, as a safeguard of data reliability, has therefore become even more important. By introducing deep autoencoder-based automatic detection, enterprises can quickly identify potential problems in massive data streams. This improves data governance efficiency, reduces risks, and strengthens the ability to respond to unexpected situations. It not only enhances the robustness of ETL processes but also supports the building of sustainable data assets[13].

In conclusion, enterprise-level ETL anomaly detection based on deep autoencoders represents both an inevitable trend of technological development and a practical requirement of digital transformation and intelligent upgrading. It enables efficient and accurate anomaly identification in complex and changing data environments. It ensures the stability and security of data processing flows and provides strong data support for enterprises in competitive markets. At the same time, this research direction promotes the development of intelligent data management systems and lays the foundation for enterprise-level big data platforms and intelligent decision-making systems. Exploring both the theoretical and practical value of this approach is therefore of great importance for achieving enterprise data-driven strategies and for generating long-term value.

## II. RELATED WORK

Early approaches to anomaly detection in ETL data streams primarily used rule-based and statistical methods, which relied on fixed thresholds, business rules, or statistical distributions to identify abnormal patterns. While effective at capturing common issues like missing values and inconsistent formats, these approaches lacked flexibility and scalability, requiring frequent manual adjustment in dynamic enterprise environments, and thus were limited in handling the increasing complexity and scale of modern data streams [14].

The evolution of machine learning introduced both supervised and unsupervised methods. Supervised models, effective with high-quality labeled anomaly data, struggled with rare anomaly occurrences and high labeling costs in enterprise-level ETL, restricting large-scale application. Unsupervised methods such as clustering and density estimation, while label-free and suitable for many scenarios, faced challenges in modeling high-dimensional, nonlinear, and multimodal features—further limiting their capacity to capture deep anomaly structures [15]. Deep learning models, including convolutional and recurrent neural networks, improved feature extraction and sequence modeling, enhancing anomaly detection accuracy and robustness. However, traditional deep learning models still encountered difficulties in generalizing to heterogeneous, non-stationary, and high-frequency ETL data and maintaining interpretability, leading to ongoing challenges in balancing model expressiveness and adaptability. To address these gaps, researchers have increasingly adopted autoencoder-based approaches, which reconstruct input data to learn latent distributions and use reconstruction errors as anomaly indicators. Autoencoders and their variants—such as sparse and variational autoencoders—have proven effective for unsupervised anomaly detection, particularly where anomaly data is scarce. Their adaptability and strong generalization ability have enabled significant progress in handling multi-source, real-time, and high-dimensional ETL data streams. As a result, autoencoder-based models now form a solid theoretical and practical foundation for developing intelligent, scalable anomaly detection systems in enterprise-level ETL processes [16].

## III. METHOD

In this study, we propose an enterprise-level ETL data stream anomaly detection method based on deep autoencoders. First, the original data stream is represented as a matrix. Let the input data be:

$$X = \{x_1, x_2, ..., x_N\}, \quad x_i \in R^d \quad (1)$$

Where N represents the number of data samples and d is the characteristic dimension of each data stream. In order to capture the multi-source heterogeneous characteristics of the data, normalization and standardization operations are required at the input stage. The normalization process can be expressed as:

$$x_i = \frac{x_i - \mu}{\sigma} \quad (2)$$

Where $\mu$ and $\sigma$ are the mean and standard deviation of the feature, respectively. This step can reduce the impact of different dimensions and ensure that the input features maintain numerical stability during training. The model architecture is shown in Figure 1.

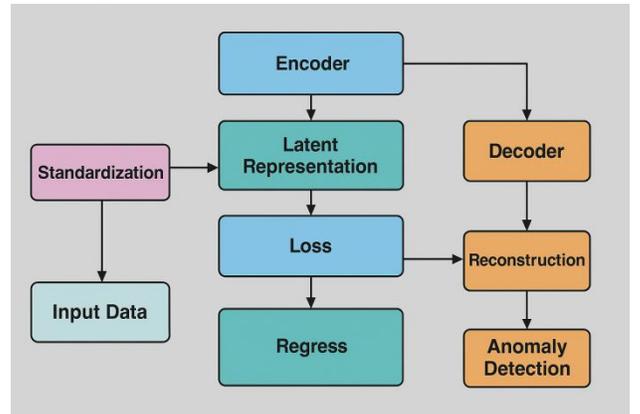

Figure 1. Overall model architecture

In the encoding stage, the autoencoder maps the input data to the potential representation space through nonlinear transformation. Let the encoding function be $f_\theta$, and its mapping process is defined as:

$$h_i = f_\theta(x_i) = \sigma(W_e \tilde{x}_i + b_e) \quad (3)$$

Where $W_e \in R^{k \times d}$ and $b_e \in R^k$ are weight and bias parameters, respectively, $\sigma(\cdot)$ represents a nonlinear activation function, and $h_i \in R^k$ is a latent vector. The encoder's goal is to map high-dimensional input to a lower-dimensional representation that preserves key features. This approach effectively captures the underlying distributional characteristics of ETL data streams and suppresses interference from noise and redundant information.

In the decoding phase, the model attempts to reconstruct the input data from the potential representation. Let the decoding function be $g_\phi$, and the reconstruction process can be expressed as:

$$x_i = g_\phi(h_i) = \sigma(W_d h_i + b_d) \quad (4)$$

Where $W_d \in R^{d \times k}, b_d \in R^d$, then the decoder restores the input through inverse mapping. If the input data is normal, the reconstruction error should be small. However, if the input data is abnormal, the reconstruction error will increase significantly because its distribution deviates from the normal pattern. Therefore, the reconstruction error becomes the core metric for measuring abnormality. This error can be defined by the mean square error (MSE):

$$L_{rec} = \frac{1}{N} \sum_{i=1}^{N} \| x_i - \widehat{x}_i \|^2 \quad (5)$$

This loss function is continuously minimized during training so that the model learns the underlying distribution of normal data.

In practical applications, to further enhance anomaly detection capabilities, we introduce regularization constraints and threshold determination mechanisms within the autoencoder framework. Regularization is implemented by imposing penalty terms on the latent representations. Drawing on the contrastive learning-based dependency modeling approach proposed by Xing et al.[17], we apply contrastive regularization to encourage the model to learn discriminative and robust latent features that are sensitive to subtle anomalies in high-dimensional ETL data.

At the same time, inspired by Feng et al.'s federated risk discrimination framework [18], we incorporate thresholding techniques that adaptively determine anomaly boundaries based on siamese network outputs, improving the accuracy and reliability of anomaly identification in distributed or privacy-sensitive settings. The penalty terms for latent space regularization can be formulated as follows:

$$L_{rec} = \lambda \| h_i \|_1 \quad (6)$$

Where $\lambda$ is the regularization coefficient, which is used to control the sparsity of the latent space and avoid overfitting. The final optimization goal combines the reconstruction error and the regularization term, and is defined as:

$$L = L_{rec} + L_{reg} \quad (7)$$

During the detection phase, when the reconstruction error $\| x_i - \widehat{x}_i \|^2$ exceeds the set threshold $\delta$, the sample is identified as an anomaly. This method enables efficient identification of complex anomalies in enterprise-level ETL data streams, ensuring data quality and system stability.

IV. EXPERIMENTAL RESULTS

A. Dataset

The dataset used in this study is derived from the ETL subset of the EDD dataset, which encompasses heterogeneous ETL processes from multiple sources. It includes multidimensional records from financial transactions, business logs, and data transmission pipelines, containing both structured and semi-structured information. This enables the simulation of complex data scenarios encountered in real enterprise operations. Each data record is indexed by a timestamp and contains raw input features, transformed fields, and loading status information, providing a complete input-output chain for building anomaly detection models.

The dataset is characterized by its large scale and high frequency, consisting of millions of data stream segments and covering dozens of feature dimensions, including transaction amount, device type, geographic location, system latency, task duration, and log markers. By jointly modeling these features, the normal operating patterns of ETL processes can be captured and potential anomalies-such as data delays, missing fields, duplicate loading, and sudden abnormal changes-can be identified. This diverse feature structure offers sufficient complexity for model training and ensures the comprehensiveness of the detection task.

In addition, the dataset provides binary labels for normal and abnormal cases. The abnormal samples encompass various types of errors, enabling the evaluation of model adaptability under different scenarios. The labels are generated based on system logs and automated verification rules, ensuring overall reliability. Because the dataset is both representative and realistic, reflecting the types of issues that may occur in enterprise-level ETL workflows, it holds significant value for validating the effectiveness and generalizability of anomaly detection methods.

B. Experimental Results

This paper first gives the results of the comparative experiment, as shown in Table 1.

Table1. Comparative experimental results

| Model | AUC | ACC | Precision | Recall |
|---|---|---|---|---|
| 1DCNN[19] | 0.914 | 0.887 | 0.862 | 0.853 |
| LSTM[20] | 0.926 | 0.892 | 0.871 | 0.861 |
| Transformer[21] | 0.938 | 0.905 | 0.884 | 0.872 |
| Mamba[22] | 0.945 | 0.913 | 0.892 | 0.879 |
| Ours | 0.963 | 0.928 | 0.911 | 0.897 |

From the results in Table 1, it can be observed that different deep learning models achieve good performance in enterprise-level ETL data stream anomaly detection tasks, but their overall performance shows some variation. Traditional sequence modeling methods such as 1DCNN and LSTM have certain advantages in feature extraction and temporal dependency modeling. However, due to their limited ability to capture nonlinear relationships and multimodal features in complex data streams, their AUC and Recall scores are relatively lower. This indicates that they still have shortcomings when handling heterogeneous enterprise data.

In contrast, the Transformer shows stronger advantages in global dependency modeling and feature interaction. Its AUC and Precision scores are significantly higher, suggesting that the self-attention mechanism can better capture hidden temporal anomaly patterns in ETL processes. This global modeling ability makes the Transformer more expressive in high-dimensional feature spaces and more suitable for complex enterprise data streams. However, the Transformer still has limitations in capturing local anomalies, which results in Recall being slightly below the optimal level. This suggests that relying only on global attention cannot fully cover the diversity of anomaly types.

The proposed method in this study achieves the best results across all four metrics, with a clear advantage in AUC and Precision. This indicates that the designed deep autoencoder architecture can effectively capture latent distribution features in ETL data streams. By combining latent space representation learning with reconstruction error discrimination, it significantly improves the accuracy and stability of anomaly detection. Compared with other methods, this model not only provides superior detection capability but also demonstrates stronger robustness. It can better adapt to the complexity and dynamic nature of enterprise applications, offering more reliable support for enterprise data governance and intelligent operations.

This paper also presents an experiment on the sensitivity of the learning rate to the single-metric AUC, and the experimental results are shown in Figure 2.

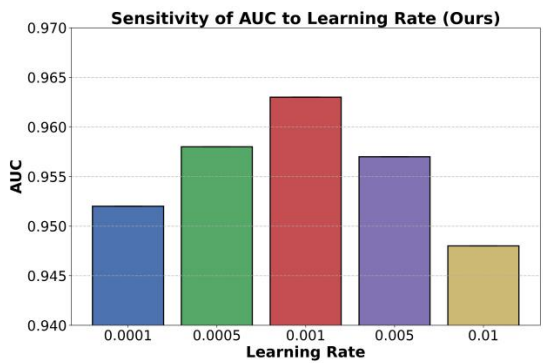

Figure 2. Sensitivity experiment of learning rate to single indicator AUC

From Figure 2, it can be observed that different learning rate settings lead to significant differences in the model's performance on the single metric AUC. When the learning rate is too low, the convergence speed of the model is limited, which results in insufficient performance and lower AUC values. This indicates that in enterprise-level ETL data stream anomaly detection tasks, an inappropriate learning rate may prevent the model from fully capturing complex feature distributions during training, thereby reducing overall detection accuracy.

As the learning rate increases to an appropriate level, the model's AUC improves significantly, reaching the best performance at a learning rate of 0.001. This shows that a moderate learning rate can achieve a good balance between convergence efficiency and stability. It allows the deep autoencoder to learn the latent representation space of normal data more effectively and to identify anomalous data streams through the reconstruction mechanism. These results highlight the critical role of hyperparameter tuning in improving the performance of anomaly detection models in enterprise applications.

When the learning rate continues to increase to a higher level, the model's performance starts to decline. At a learning rate of 0.01, the AUC drops sharply. This indicates that an excessively high learning rate can cause unstable gradient updates during training, making it difficult for the model to approach the optimal solution. Such instability is especially pronounced in complex data stream environments, where non-stationarity and high-dimensional features further amplify fluctuations in training, reducing the model's ability to recognize anomaly patterns.

Overall, the experimental results reveal the sensitivity of deep autoencoder-based anomaly detection methods to the learning rate. A moderate learning rate not only ensures efficient convergence but also enhances the model's ability to capture anomaly signals in heterogeneous data. This helps maintain the reliability and stability of detection systems in enterprise-level ETL scenarios. It also suggests that fine-tuning the learning rate is a key step to achieve optimal model performance in practice.

This paper also presents an experiment on the sensitivity of the latent dimension size to the single-metric precision. The experimental results are shown in Figure 3.

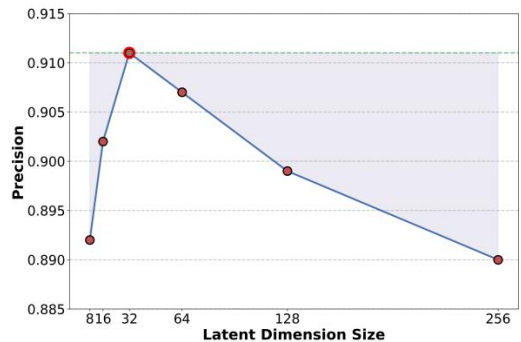

Figure 3. Sensitivity experiment of latent dimension size to single-index precision

From Figure 3, it can be observed that the size of the latent dimension shows clear sensitivity in terms of Precision. When the latent dimension increases from smaller values to a moderate scale, such as 32 dimensions, the model achieves its best Precision. This indicates that with an appropriate latent size, the autoencoder can effectively capture the underlying feature distribution of ETL data streams. At the same time, it avoids interference from redundant information and noise, making the anomaly detection process more accurate and reliable. However, as the latent dimension continues to increase,

Precision shows a gradual decline. This suggests that an excessively large latent space may cause overfitting or redundancy in feature representations. As a result, the ability to distinguish between normal and anomalous samples is weakened. In enterprise-level ETL scenarios, maintaining a moderate latent dimension is beneficial for achieving both accuracy and robustness in detection. Therefore, selecting a reasonable dimension size is a key factor in building efficient anomaly detection models.

## V. Conclusion

This study focuses on anomaly detection in enterprise-level ETL data streams and proposes a detection method based on deep autoencoders. By building an end-to-end representation learning and reconstruction mechanism, the model can automatically capture latent distribution features in heterogeneous data and identify anomalies precisely through reconstruction error. Compared with traditional rule-based or statistical methods, this approach shows clear advantages in handling complex environments and high-dimensional features. It provides strong technical support for ensuring the stability and reliability of enterprise data processing.

During the research, the method considers both feature standardization and effective latent space modeling, while introducing regularization constraints in the optimization objective to enhance robustness. This design allows better adaptation to various anomaly types in practical scenarios, including delays, missing values, duplicate loading, and sudden abnormal changes. The results demonstrate that the method achieves high detection accuracy while reducing the risk of false positives and false negatives, showing broad applicability in enterprise environments. From the perspective of application value, the proposed method is not only a technical improvement in anomaly detection but also has significant impact on data governance, intelligent operations, and real-time decision-making. This is particularly important for industries such as finance, healthcare, telecommunications, and retail that rely on real-time data. It helps improve business continuity, reduce economic losses, and provide a high-quality foundation for further data analysis and intelligent applications.

Future research can expand in several directions. One direction is to integrate the method with distributed computing frameworks to support real-time detection in ultra-large-scale data streams. Another is to introduce cross-modal data and multi-task learning mechanisms to enhance the comprehensiveness and interpretability of anomaly detection. In addition, further work can be carried out on model interpretability, adaptation to concept drift, and coordination with automated decision systems, promoting the development of enterprise-level anomaly detection toward greater intelligence and practicality.